\documentclass[letterpaper]{article} 
\usepackage{aaai2026}  
\usepackage{times}  
\usepackage{helvet}  
\usepackage{courier}  
\usepackage[hyphens]{url}  
\usepackage{graphicx} 
\usepackage{natbib}  
\usepackage{caption} 
\usepackage{algorithm}
\usepackage{algorithmic}

\usepackage[utf8]{inputenc} 
\usepackage{url}            
\usepackage{booktabs}       
\usepackage[table]{xcolor}
\usepackage{amsfonts}       
\usepackage{nicefrac}       
\usepackage{microtype}      
\usepackage{xcolor}         
\usepackage{multirow}
\usepackage{makecell}
\usepackage{amsmath}
\usepackage{colortbl}
\usepackage{rotating}
\usepackage{pifont}
\usepackage{subcaption}

\usepackage{newfloat}
\usepackage{listings}
\DeclareCaptionStyle{ruled}{labelfont=normalfont,labelsep=colon,strut=off} 
\floatstyle{ruled}
\newfloat{listing}{tb}{lst}{}
\floatname{listing}{Listing}
\title{Stratified Knowledge-Density Super-Network for Scalable Vision Transformers}
\author {
    Longhua Li\textsuperscript{\rm 1,\rm 2},
    Lei Qi\textsuperscript{\rm 1,\rm 2\thanks{Co-corresponding authors.}},
    Xin Geng\textsuperscript{\rm 1,\rm 2\footnotemark[1]}
}
\affiliations {
    \textsuperscript{\rm 1}School of Computer Science and Engineering, Southeast University, Nanjing 210096, China\\
    \textsuperscript{\rm 2}Key Laboratory of New Generation Artificial Intelligence Technology and Its Interdisciplinary\\Applications (Southeast University), Ministry of Education, China\\
    \{lhli, qilei, xgeng\}@seu.edu.cn
}

\usepackage{bibentry}

\begin{document}

\maketitle

\begin{abstract}
Training and deploying multiple vision transformer (ViT) models for different resource constraints is costly and inefficient.
To address this, we propose transforming a pre-trained ViT into a stratified knowledge-density super-network, where knowledge is hierarchically organized across weights. This enables flexible extraction of sub-networks that retain maximal knowledge for varying model sizes.
We introduce \textbf{W}eighted \textbf{P}CA for \textbf{A}ttention \textbf{C}ontraction (WPAC), which concentrates knowledge into a compact set of critical weights. WPAC applies token-wise weighted principal component analysis to intermediate features and injects the resulting transformation and inverse matrices into adjacent layers, preserving the original network function while enhancing knowledge compactness.
To further promote stratified knowledge organization, we propose \textbf{P}rogressive \textbf{I}mportance-\textbf{A}ware \textbf{D}ropout (PIAD). PIAD progressively evaluates the importance of weight groups, updates an importance-aware dropout list, and trains the super-network under this dropout regime to promote knowledge stratification.
Experiments demonstrate that WPAC outperforms existing pruning criteria in knowledge concentration, and the combination with PIAD offers a strong alternative to state-of-the-art model compression and model expansion methods.
\end{abstract}


\section{Introduction}
Vision Transformers (ViTs) \cite{dosovitskiy2020image} have demonstrated remarkable performance across various computer vision tasks \cite{radford2021learning,jose2025dinov2,hua2024reconboost,huaopenworldauc}. However, deploying ViTs in real-world scenarios often requires adapting model size to diverse resource constraints, ranging from edge devices with limited memory and compute to powerful servers requiring high-throughput inference. The conventional approach of training and maintaining multiple ViT variants for each deployment setting is computationally expensive and inefficient. Moreover, many existing pruning or model compression methods either require extensive fine-tuning or suffer from performance degradation due to suboptimal knowledge retention.

Recent works \cite{xia2024transformer, xia2024initializing, xia2024exploring, feng2025wave} have adopted the Learngene paradigm, which extracts or distills core parameters from a pre-trained model and expands them to construct models of varying sizes. However, these methods rely on manually designed expansion rules to reuse a limited portion of refined weights, making it difficult to introduce new or size-specific knowledge during expansion, thus limiting the capacity alignment of the resulting models.

\begin{figure}[t]
    \centering
    \includegraphics[width=\linewidth]{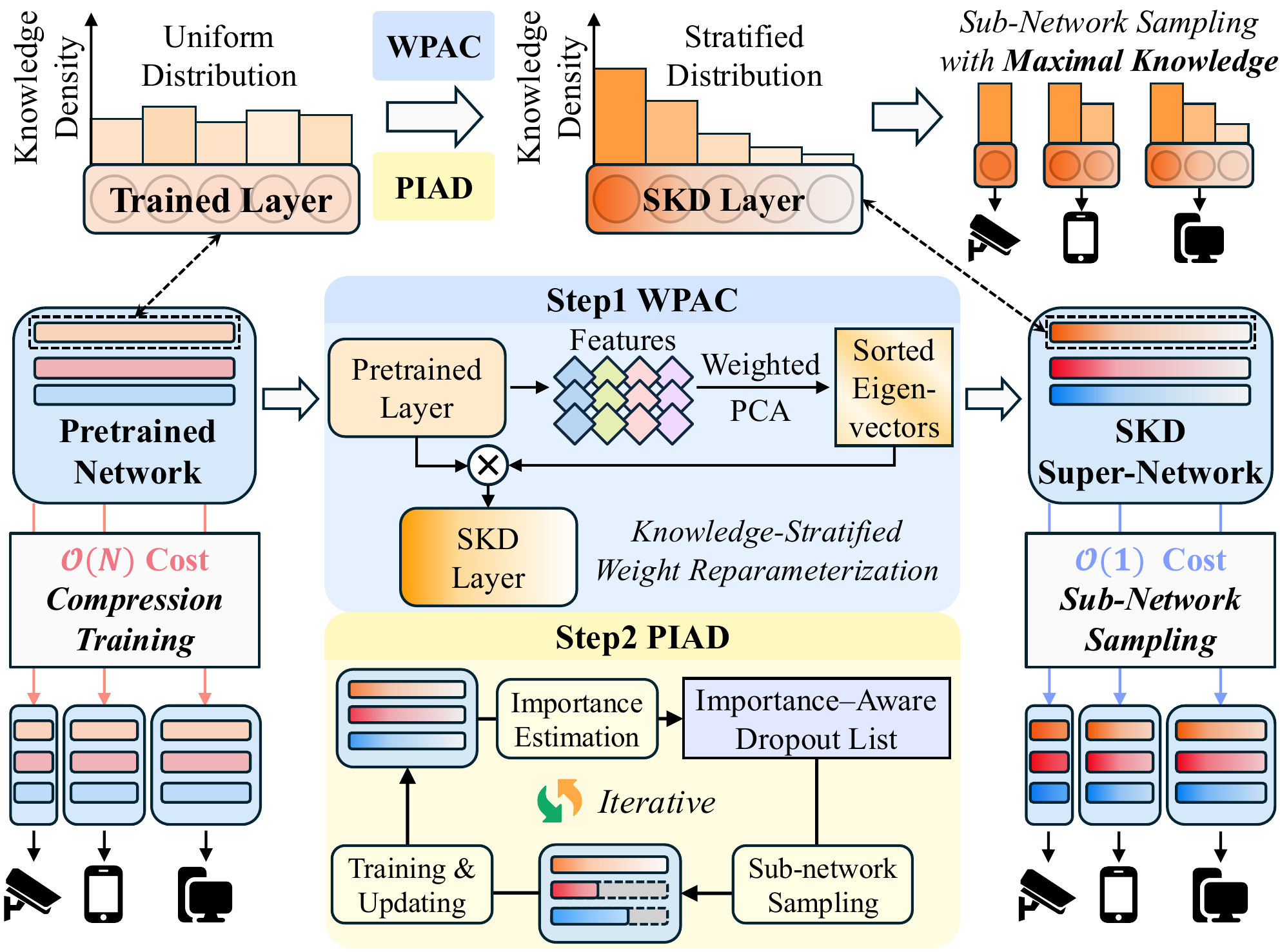}
    \caption{An overview of the proposed method. Traditional methods require separate compression for each deployment setting. In contrast, our method builds a \textbf{S}tratified \textbf{K}nowledge-\textbf{D}ensity (SKD) Super-Network in a single pass, enabling cost-free sub-network extraction for arbitrary model sizes.}
    \label{fig:overview}
\end{figure}

To address these challenges, we propose transforming a pre-trained ViT into a \textbf{S}tratified \textbf{K}nowledge-\textbf{D}ensity (SKD) Super-Network, as illustrated in Figure \ref{fig:overview}. Guided by knowledge density, sub-networks of various sizes are extracted to maximize knowledge retention relative to their capacity.

Specifically, we introduce WPAC to perform initial stratification of knowledge density. WPAC leverages a small proxy dataset, randomly sampled from the training set, to compute intermediate attention features. These features are weighted token-wise using Taylor-based importance scores and then decomposed via PCA to obtain a projection matrix composed of ordered principal components. The projection and its inverse are integrated into the layers surrounding the intermediate features, preserving the original network function while concentrating knowledge into a smaller set of critical dimensions, thus achieving a stratified distribution of knowledge density across parameter dimensions.

To further enhance this stratification, we propose PIAD, which assigns lower dropout probabilities to more important parameter groups and higher probabilities to less important ones. This progressively amplifies the distinction in knowledge density, particularly improving the performance of smaller sub-networks.
Together, WPAC and PIAD enable a single ViT model to function as a scalable backbone, from which sub-networks of arbitrary sizes can be extracted at $\mathcal{O}(1)$ cost based on the learned importance ranking.

Our contributions can be summarized as follows:
\begin{itemize}
  \item We propose WPAC, a weighted PCA mechanism that condenses knowledge into fewer dimensions via function-preserving transformations. It consistently outperforms existing pruning criteria in knowledge concentration.
  \item We introduce PIAD, a progressive dropout mechanism that builds a scalable super-network via importance-aware sub-network sampling and training. It achieves flexible control over compression while maintaining accuracy.
  \item We achieve strong results on standard benchmarks, outperforming or matching state-of-the-art model pruning and scaling methods. Our framework enables flexible and efficient ViT deployment across diverse resource constraints.
\end{itemize}

\section{Related Work}

\paragraph{Network Pruning.}
Network pruning \cite{lecun1989optimal,fang2023depgraph} reduces model complexity by removing redundant or less important weights.
While magnitude-based pruning \cite{han2015learning,li2018optimization,lee2021layer} is widely used, small weights can still play critical roles.
FPGM \cite{he2019filter} leverages geometric medians to identify redundant filters. Other methods construct importance metrics based on activation values \cite{hu2016network,muralidharan2024compact} or information entropy \cite{luo2017entropy}, while many approaches \cite{molchanov2017pruning,molchanov2019importance} apply Taylor expansion to approximate the loss induced by pruning.
However, these methods often require time-consuming pruning for each target size.

\paragraph{Scalable Neural Networks.}
Scalable neural networks are designed to support flexible model capacity adjustment for different resource constraints. Notable examples include Slimmable Networks \cite{yu2019slimmable} and Once-for-All \cite{cai2020once}, which enables sub-network extraction with elastic size.
Recently, a paradigm called Learngene \cite{xia2024transformer,xia2024exploring,feng2025wave, li2025one} was proposed, which distills core weights and expands them to generate descendant models.
In contrast, we introduce a stratified knowledge-density structure to enable fast extraction of sub-networks with maximal knowledge across sizes.

\paragraph{Low-Rank Compression.}
Low-rank approximation is a widely used strategy for compressing deep models by exploiting the redundancy in weight matrices. Techniques such as SVD and other matrix factorization methods have been applied to reduce parameter count and computational cost \cite{noach2020compressing,wang2020structured,mao2020ladabert,yu2023compressing}.
Instead of splitting a layer into two low-rank components, we perform PCA on intermediate features and apply the transformation and its inverse to adjacent layers, concentrating knowledge into a few dimensions while preserving network function.

\section{Method}
Figure \ref{fig:overview} shows the framework of our method. WPAC first uses PCA to induce a stratified distribution of knowledge in the pre-trained model, and PIAD further enhances this effect, resulting in a stratified knowledge-density super-network. This enables cost-free extraction of sub-networks that retain maximal knowledge, providing strong initialization for downstream scenarios with varying resource constraints.

\subsection{Preliminaries}
We consider a standard ViT consisting of a stack of transformer blocks, each with a Multi-Head Self-Attention (MHSA) module followed by a feed-forward network (MLP).

In MHSA, each attention head uses four linear projections: $W_q, W_k, W_v, W_o \in \mathbb{R}^{d \times d}$, with corresponding biases $b_q, b_k, b_v, b_o \in \mathbb{R}^{d \times 1}$.
The MLP consists of two fully connected layers with a hidden dimension of $d^{\prime}$. The weight and bias of the first and second linear layers are denoted as $W_{1}\in\mathbb{R}^{d^{\prime}\times d},b_{1}\in\mathbb{R}^{d^{\prime} \times 1}$, and $W_{2}\in\mathbb{R}^{d\times d^{\prime}},b_{2}\in\mathbb{R}^{d \times 1}$.

\begin{figure*}[t]
    \centering
    \includegraphics[width=\linewidth]{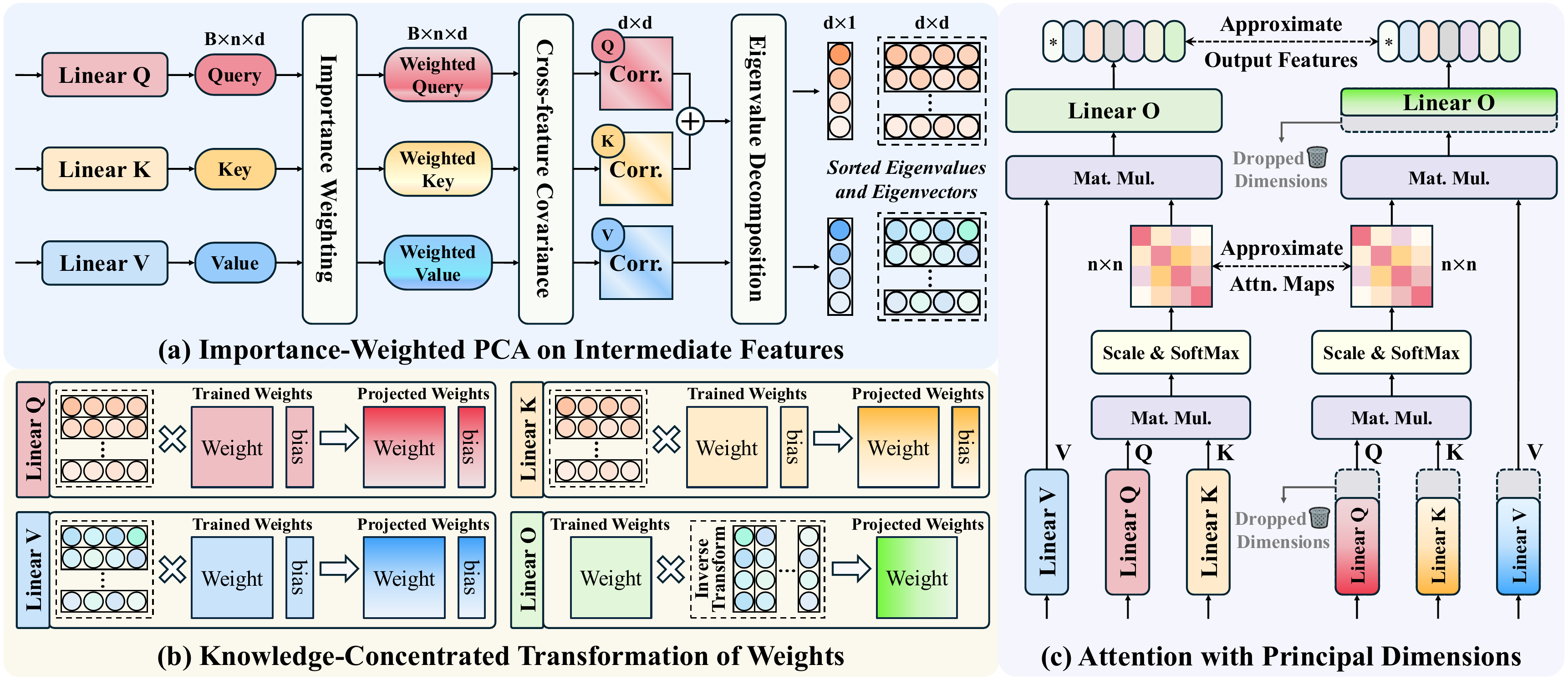}
    \caption{An overview of the proposed WPAC. (a) Principal component projection matrices are computed using importance-weighted intermediate features. (b) The resulting transformation matrices are applied to the pre-trained weights. (c) The transformed linear layers can preserve the original neural function using only a small number of principal dimensions.}
    \label{fig:wpac}
\end{figure*}

\begin{figure*}[t]
    \centering
    \includegraphics[width=\linewidth]{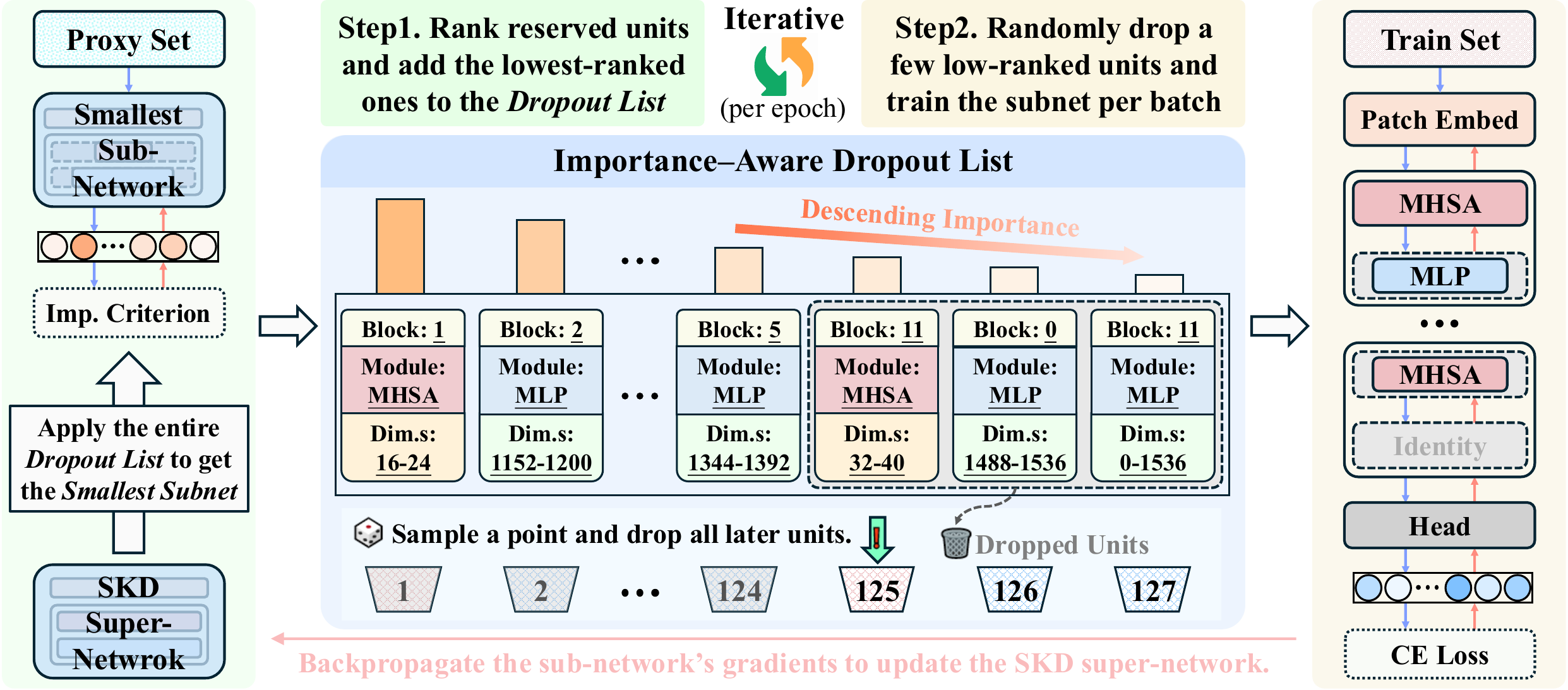}
    \caption{An overview of the proposed PIAD. \textbf{Step 1.} At the start of each epoch, evaluate the importance of parameter groups not yet in the dropout list and append the least important ones. \textbf{Step 2.} During training, sample sub-networks by randomly dropping the least important units from the dropout list and train them, propagating gradients back to the \textit{SKD Super-Netwrok}.}
    \label{fig:piad}
\end{figure*}

\subsection{Weighted PCA for Attention Concentration}
Figure \ref{fig:wpac} illustrates the WPAC method, which uses a small proxy dataset $\mathcal{D}$ and PCA to transform the pre-trained weight matrix, concentrating knowledge into key dimensions. This transformation is applied separately to each head. For brevity, head indices are omitted unless otherwise specified.

\paragraph{Transformation for Linear Projections V and O.}
Let ${X_{v}\in\mathbb{R}^{n\times d}}$ denote the output features of the value projection layer, where $n$ is the number of tokens. We center the data and compute its covariance matrix $\mathbf{Corr}\in\mathbb{R}^{{d}\times{d}}$. This covariance is updated incrementally across the proxy set. Applying eigen decomposition to $\mathbf{Corr}$ yields eigenvalues and eigenvectors, sorted in descending order. The eigenvectors form a transformation matrix ${W_{\mathrm{Trans}}^{(vo)}}{\in}\mathbb{R}^{{d}\times{d}}$, where each row corresponds to a principal component.
We apply $W_{\mathrm{Trans}}^{(vo)}$ to the value and output projections as follows:
\begin{equation}
    {W_{v}}\leftarrow{W_{\text{Trans}}^{(vo)}}{W_{v}},\ {b_{v}}\leftarrow{W_{\text{Trans}}^{(vo)}}{b_{v}},\ {W_{o}}\leftarrow{W_{o}}{(W_{\text{Trans}}^{(vo)})^{-1}}.
\end{equation}
This transformation satisfies:
\begin{itemize}
    \item \textbf{Function Preservation}: Let ${x\in\mathbb{R}^{d}}$ be a token representation input to the attention mechanism. Ignoring the attention map's weighting across tokens, the original forward pass is $W_{o} \left( W_{v}x+b_{v} \right)+b_{o}$. After applying the WPAC transformation, the equivalent form becomes ${W}_{o}({W}_{\text{Trans}}^{(vo)})^{-1}({W}_{\text{Trans}}^{(vo)}{W}_{v}{x}+ {W}_{\text{Trans}}^{(vo)}{b}_{v})+{b}_{o}$, which remains mathematically equivalent to the original.
    \item \textbf{Information Concentration}: Let $f_{v}=W_{v}x+b_{v}$ denote the output of the linear projection V. By extracting the top-$k$ principal components, we obtain $({W}_{{\mathrm{Trans}}}^{(vo)})^{[:k,:]}f_v$, which retains most of the original feature’s information. This is equivalent to retaining only the top-$k$ dimensions of the transformed linear projection parameters:
    \begin{equation}({W}_{{\mathrm{Trans}}}^{(vo)})^{[:k,:]}f_v\equiv({W}_{\text{Trans}}^{(vo)}{W}_{v})^{[:k,:]}{x}+ ({W}_{\text{Trans}}^{(vo)}{b}_{v})^{[:k]}.
    \end{equation}
\end{itemize}

\paragraph{Transformation for Linear Projections Q and K.}

For the query and key projections, we compute the covariance matrices of their output token representations and sum them before applying eigen decomposition. The resulting transformation matrix ${W}^{(qk)}_{{\mathrm{Trans}}}$ is then applied as follows:
\begin{equation}
    \begin{split}
    {W_{q}}\leftarrow{W_{\text{Trans}}^{(qk)}}{W_{q}},\quad{b_{q }}\leftarrow{W_{\text{Trans}}^{(qk)}}{b_{q}},\\{W_{k}}\leftarrow{W_{\text{Trans}}^{(qk)}}{W_{k}},\quad{b_{k }}\leftarrow{W_{\text{Trans}}^{(qk)}}{b_{k}}.
    \end{split}
\end{equation}
This transformation ensures:
\begin{itemize}
    \item \textbf{Function Preservation}: The original similarity between input tokens $x_i$ and $x_j$ is computed as follows:
    \begin{equation}
    \text{sim}_{(i,j)}=\left( W_{q}x_i+b_{q} \right)^{T} \left( W_{k}x_j+b_{k} \right).
    \end{equation}
    Using the transformed projections, the similarity becomes $(W_{\text{Trans}}^{(qk)}W_{q}{x_i}+W_{\text{Trans}}^{(qk)}b_{q})^{T}(W_{\text{Trans}}^{(qk)}W_{k} {x_j}+W_{\text{Trans}}^{(qk)}b_{k})$.
    Through algebraic manipulation, this can be rewritten as $(W_{q}x_i+b_{q})^{T}(W_{\text{Trans}}^{(qk)})^TW^{(qk)}_{\text{Trans}} \left( W_{k}x_j+b_{k} \right)$.
    Since $W_{\text{Trans}}^{(qk)}$ is orthogonal, we obtain the following identity:
    \begin{equation}
    \text{sim}_{(i,j)}\equiv\left( W_{q}x_i+b_{q} \right)^{T}(W_{\text{Trans}}^{(qk)})^TW^{(qk)}_{\text{Trans}} \left( W_{k}x_j+b_{k} \right).
    \end{equation}
    \item \textbf{Information Concentration}: Let $\boldsymbol{q_i}=W_{q}x_i+b_{q}$ and $\boldsymbol{k_j}=W_{k}x_j+b_{k}$. When retaining only the top-$k$ dimensions of the transformed linear projections Q and K, the resulting similarity score between tokens $x_i$ and $x_j$ is:
    \begin{equation}\text{sim}^{\text{top-}k}_{(i,j)}=\boldsymbol{q_i}((W_{\text{Trans}}^{(qk)})^{[:k,:]})^T(W_{\text{Trans}}^{(qk)})^{[:k,:]}\boldsymbol{k_j}.
    \end{equation}
    Since PCA preserves principal components of the original features, the following approximation holds:
    \begin{equation}\boldsymbol{k_j} \approx ((W_{\text{Trans}}^{(qk)})^{[:k,:]})^T(W_{\text{Trans}}^{(qk)})^{[:k,:]}\boldsymbol{k_j},
    \end{equation}
    which leads to:
    \begin{equation}
    \text{sim}^{\text{top-}k}_{(i,j)} \approx \boldsymbol{q_i}^{T} \boldsymbol{k_j} = \text{sim}_{(i,j)}.
    \end{equation}
    Therefore, keeping only the top-$k$ dimensions of the transformed linear projections yields an approximate similarity matrix that closely matches the original.
\end{itemize}

\paragraph{Weighted PCA.}
\label{sec:weighted_pca}
PCA inherently treats all tokens and feature dimensions equally, which may not reflect their true contributions to prediction. To address this, we introduce token-wise weighting based on first-order Taylor importance.
The importance of a parameter or output element $h_i$ is estimated by the change in cost $\mathcal{C}(\mathcal{D})$, evaluated on a proxy set $\mathcal{D}$, when $h_i$ is set to zero. The estimated importance score is:
\begin{equation}
    \Theta_{\text{TE}}(h_{i})=\big|\Delta\mathcal{C}(h_{i})\big|\approx\bigg|\frac{\delta\mathcal{C}}{\delta h_{i}}\cdot h_{i} \bigg|.
\end{equation}
Concretely, we compute the element-wise importance $\Theta_{\text{TE}}\in\mathbb{R}^{n\times d}$ for intermediate token representations $X\in\mathbb{R}^{n\times d}$in the attention module, where $n$ is the number of tokens and $d$ is the feature dimension. We then aggregate over the feature dimension to obtain token-wise importance scores $\Theta^{\text{token}}_{TE}\in\mathbb{R}^{n}$.
We apply a square-root weighting to the centered token features $\tilde{X}$ before computing the covariance matrix:
\begin{equation}
    \tilde{X}\leftarrow\text{diag}(\sqrt{\Theta_{\text{TE}}^{\text{token}}})\cdot {\tilde{X}},\quad\mathbf{Corr}=\frac{1}{n-1}{\tilde{X}}^{T}{\tilde{X}}.
\end{equation}
This transformation biases the resulting principal components to better preserve high-importance token directions, while maintaining stability during eigen decomposition.

\paragraph{Transformation for MLP.}
\label{seq:MLPtrans}
Due to the non-linear activation between MLP layers, applying PCA directly may distort the encoded information.
As an alternative, we rank dimensions using Taylor importance. We first compute element-wise scores $\Theta_{\text{TE}}\in\mathbb{R}^{n\times d}$, then aggregate across tokens to obtain dimension-wise importance $\Theta^{\text{dim}}_{\text{TE}}\in\mathbb{R}^{d}$. A sorting matrix $W_{\text{sort}}$ is then constructed to reorder dimensions accordingly:
\begin{equation}
    W_{\text{sort}}[i,j]=\begin{cases}1,&\text{if the }j\text{-th dimension ranks }i\text{-th,}\\ 0,&\text{otherwise.}\end{cases}
\end{equation}
We then apply the transformation and its inverse to the MLP:
\begin{equation}
    W_1\leftarrow W_{\text{sort}}W_1,\ b_1\leftarrow W_{\text{sort}}b_1,\ W_2\leftarrow W_2W^{-1}_{\text{sort}}.
\end{equation}
This process empirically preserves the original function while concentrating information into the top-ranked dimensions.

\subsection{Progressive Importance-Aware Dropout}
Figure \ref{fig:piad} shows PIAD for enhancing knowledge stratification.
After applying the above neural-equivalent transformation for knowledge concentration, the resulting network—referred to as the \textit{SKD Network}—possesses a degree of scalability.
To further enhance its adaptability across different model sizes, we introduce Progressive Importance-Aware Dropout (PIAD), a two-stage process repeated in each training epoch:
\begin{itemize}
    \item \textbf{Before each epoch:} Rank reserved compression units and add the lowest-importance ones to the \textit{Dropout List}.
    \item \textbf{During each epoch:} In each iteration, randomly drop several low-importance units in the dropout list, train the sub-network, and backpropagate to the \textit{SKD Network}.
\end{itemize}

\paragraph{Building and Evaluating Droppable Units.}
The dropout list contains candidate droppable units, each assigned an importance score for ranking. Specifically, we divide the intermediate dimensions of each MHSA into 8 groups and those of each MLP into 32 groups, with each group treated as a droppable unit. Each group forms one droppable unit.
To evaluate importance, we randomly sample a small proxy set $\mathcal{D}$ from the training data and proceed in two steps:
\begin{itemize}
    \item \textbf{Module Sensitivity}: For each module $m$ (e.g., MHSA or MLP), we define its sensitivity $\gamma_m$ as the relative increase in the cost $\mathcal{C}( \mathcal{D})$ when the module is skipped:
    \begin{equation}
        \gamma_{m}=\frac{\mathcal{C}(\mathcal{D}\mid m\ \text{skipped})-\mathcal{C}(\mathcal{D})}{\mathcal{C}( \mathcal{D})}.
    \end{equation}
    \item \textbf{Dimension Importance}: For each dimension $i$ in module $m$, we compute its importance score $\Theta_{i}^{ \left( m \right)}$ using first-order Taylor expansion. These scores are normalized within the module to obtain contribution ratios $\alpha_{i}^{(m)}$, and the final global importance is defined as $I_{i}^{(m)}$:
    \begin{equation}
    \alpha_{i}^{(m)}=\frac{{\Theta_{i}^{(m)}}}{\sum_{{j}\in{m}}{\Theta_{j}^{(m)}}},\quad I_{i}^{ \left( m \right)}= \gamma_{m} \cdot \alpha_{i}^{ \left( m \right)}.
    \end{equation}
\end{itemize}
Each droppable unit $u$ (i.e., a group of dimensions) accumulates the scores of all its constituent dimensions:
\begin{equation}
    I_{u}= \frac{ \sum_{i \in u}I_{i}^{ \left( m \right)}}{\text{MACs} \left( u \right)},
\end{equation}
where $\text{MACs}(u)$ denotes the MACs of unit $u$. These final scores $I_{u}$ are used to rank the units for dropout selection.

\begin{table*}[t]
\setlength{\tabcolsep}{3.05pt}
\begin{tabular}{lccccccccccccccccc}
\toprule[1.5pt] 
\multirow{2}{*}{Method} & \multirow{2}{*}{Epochs} & \multirow{2}{*}{KD} & \multicolumn{5}{c}{DeiT-B (81.8)} & \multicolumn{5}{c}{DeiT-S (79.8)} & \multicolumn{5}{c}{DeiT-Ti (72.1)}
\\
\cmidrule[0.5pt](lr){4-8}\cmidrule[0.5pt](lr){9-13}\cmidrule[0.5pt](lr){14-18} &&& 4:12 & 6:12 & 8:12 & 10:12 & 12:12 & 4:12 & 6:12 & 8:12 & 10:12 & 12:12 & 4:12 & 6:12 & 8:12 & 10:12 & 12:12
\\
\midrule[0.5pt] 

Albert & 10/10/10 & \textit{w/o} & 71.7 & 75.3 & 76.4 & 77.4 & 77.6 & 65.0 & 69.7 & 71.7 & 72.7 & 73.3 & 55.2 & 59.8 & 62.5 & 64.3 & 65.3 \\
LiGO & 10/10/10 & \textit{w/o} & - & 74.2 & 74.4 & 75.3 & 75.4 & - & 68.6 & 69.9 & 69.7 & 70.0 & - & 59.0 & 60.2 & 59.8 & 60.9 \\
Heur-LG & 10/10/10 & \textit{w/o} & 60.5 & 68.7 & 72.2 & 73.6 & 74.0 & 52.3 & 57.3 & 61.7 & 64.4 & 65.9 & 41.5 & 47.4 & 50.5 & 53.5 & 55.5 \\
Auto-LG & 10/10/10 & \textit{w/o} & 60.9 & 70.0 & 72.4 & 73.5 & 73.8 & 63.2 & 70.5 & 72.2 & 73.3 & 73.8 & 52.4 & 61.8 & 64.6 & 65.9 & 66.8 \\
TLEG & 40/35/50 & \textit{w} & 71.6 & 76.2 & 78.1 & 79.1 & 79.9 & 63.7 & 69.5 & 72.3 & 73.9 & 75.1 & - & 58.2 & - & - & 65.4 \\
SWS & 10/10/10 & \textit{w} & - & 76.9 & 79.0 & 79.7 & 80.1 & - & 70.2 & 73.4 & 75.1 & 75.8 &-&-&-&-&- \\
WAVE & 10/10/10 & \textit{w} & 74.5 & 77.5 & 78.2 & 78.9 & 79.2 & 68.9 & 72.7 & 74.1 & 74.9 & 75.3 & 58.6 & 63.2 & 65.4 & 66.6 & 67.3 \\
\rowcolor[HTML]{C6E2FF}SKD (Ours) & 0/0/0 & \textit{w/o} & \textbf{77.0} & \textbf{80.4} & \textbf{80.9} & \textbf{81.5} & \textbf{81.5} & \textbf{70.6} & \textbf{76.2} & \textbf{78.2} & \textbf{79.0} & \textbf{79.0} & \textbf{61.4} & \textbf{65.8} & \textbf{68.6} & \textbf{70.0} & \textbf{70.8} \\


\bottomrule[1.5pt] 
\end{tabular}
\caption{Performance comparison with network expansion methods on ImageNet-1k. ``$x$:12'' denotes the ratio of the scaled model’s size to the original model. ``KD'' indicates whether additional knowledge distillation is required during the construction of the scalable network. ``Epochs'' refers to the number of fine-tuning epochs used for the scaled DeiT-B/S/Ti models.}
\label{tab:network_expansion}
\end{table*}

\paragraph{Progressively Updating Dropout List.}
We adopt a progressive schedule to update the dropout list. Let the target maximum compression ratio be $r$, and define the progressive phase to span $P_e$ epochs. At the beginning of epoch $t$, we select additional droppable units from the remaining candidates such that their total MACs approximate $\frac{r}{P_e}\times \text{MACs}_O$, where $\text{MACs}_O$ denotes the MACs of the \textit{SKD Network}. Specifically, we first form the \textit{Smallest Sub-Network} by dropping all units currently in the list, then evaluate the importance of the remaining units using the proxy set.
While traversing units from high to low dimensions within a module, we merge any unit with lower importance than the highest previously seen into that unit, and average their importance scores.
We continue appending the lowest-ranked units to the dropout list until its accumulated MACs reach $\frac{t\times r}{P_e}\times \text{MACs}_O$.

\paragraph{Sampling and Training Sub-networks.}
The updated dropout list is applied throughout the subsequent training epoch. Specifically, for each batch, we uniformly sample a truncation index $s$ from 1 to the current length of the dropout list. All low-importance units ranked beyond the $s$-th position are dropped, forming a sampled sub-network. We then perform forward and backward passes on this sub-network and propagate the gradients back to the \textit{SKD Network}.

\paragraph{}
This cycle lasts for $P_e$ epochs, after which the dropout list is no longer updated. In the subsequent epochs, sub-networks continue to be sampled and trained, and the \textit{SKD Network} is updated accordingly.
Eventually, the \textit{SKD Network} evolves into a scalable super-network. At deployment, a model of the desired size can be instantiated by discarding the low-ranked dropout units until the target MACs is reached.


\section{Experiments}

\subsection{Experimental Setup}
We conduct experiments primarily on the DeiT \cite{touvron2021training} models, as well as Swin Transformers \cite{liu2021swin} pretrained on ImageNet-1k \cite{russakovsky2015imagenet}. During the WPAC and PIAD stages, we randomly sample tiny training sets of size 1024 from ImageNet-1k to serve as proxy sets.
We report the top-1 accuracy using a fixed random seed of 0. When reporting the standard deviation, we repeat each experiment 5 times using random seeds from 0 to 4.
All experiments across datasets are conducted at a resolution of $224 \times 224$. Unless otherwise specified, our training settings follow those of DeiT.
The \textit{SKD Network} is obtained by applying WPAC to the DeiT pretrained model. During the subsequent PIAD stage, DeiT-B is trained for 150 epochs, and DeiT-S and DeiT-Ti are trained for 300 epochs to construct the \textit{SKD Network}. We set $P_e=50$ as the number of epochs for progressively constructing the dropout list.

\subsection{Main Results}

\begin{table}[t]
\setlength{\tabcolsep}{3.05pt}
\begin{tabular}{llcccc}
\toprule[1.5pt] 
Backbone & Method & MACs & Params & Epochs & Top-1
\\
\midrule[0.5pt] 

\multirow{12}{*}{DeiT-S} & \color{gray}Original & \color{gray}4.26 G & \color{gray}22.05 M & \color{gray}- & \color{gray}79.83 \\
 & IA-RED$^2$ & 3.60 G & - & 90 & 79.30 \\ 
 & IA-RED$^2$ & 3.30 G & - & 90 & 78.40 \\ 
 & SPViT & 3.30 G & 15.90 M & 300 & 78.30 \\ 
 & RePaViT & 3.20 G & 16.70 M & 300 & 78.90 \\ 
 & WDPruning & 3.10 G & 15.00 M & 100 & 78.55 \\
 & \cellcolor[HTML]{C6E2FF}SKD (Ours) & \cellcolor[HTML]{C6E2FF}3.07 G & \cellcolor[HTML]{C6E2FF}16.03 M & \cellcolor[HTML]{C6E2FF}30 & \cellcolor[HTML]{C6E2FF}\textbf{79.42} \\

 & IA-RED$^2$ & 2.90 G & - & 90 & 78.60 \\ 
 & WDPruning & 2.60 G & 13.30 M & 100 & 78.38 \\
 & RePaViT & 2.50 G & 13.20 M & 300 & 77.00 \\ 
 & CP-ViT & 2.46 G & - & 30 & \textbf{79.08} \\
 & \cellcolor[HTML]{C6E2FF}SKD (Ours) & \cellcolor[HTML]{C6E2FF}2.43 G & \cellcolor[HTML]{C6E2FF}12.78 M & \cellcolor[HTML]{C6E2FF}30 & \cellcolor[HTML]{C6E2FF}78.71 \\

\midrule[0.5pt] 


\multirow{6}{*}{DeiT-Ti} & \color{gray}Original & \color{gray}1.08 G & \color{gray}5.72 M & \color{gray}- & \color{gray}72.14 \\
 & SPViT & 1.00 G & 4.80 M & 300 & 70.70 \\ 
 & WDPruning & 0.90 G & 3.80 M & 100 & 71.10 \\
 & \cellcolor[HTML]{C6E2FF}SKD (Ours) & \cellcolor[HTML]{C6E2FF}0.89 G & \cellcolor[HTML]{C6E2FF}4.77 M & \cellcolor[HTML]{C6E2FF}30 & \cellcolor[HTML]{C6E2FF}\textbf{71.40} \\
 & RePaViT & 0.80 G & 4.40 M & 300 & 69.40 \\ 
 & \cellcolor[HTML]{C6E2FF}SKD (Ours) & \cellcolor[HTML]{C6E2FF}0.79 G & \cellcolor[HTML]{C6E2FF}4.25 M & \cellcolor[HTML]{C6E2FF}30 & \cellcolor[HTML]{C6E2FF}\textbf{70.46} \\

\bottomrule[1.5pt] 
\end{tabular}
\caption{Comparison with network compression methods on ImageNet-1k based on DeiT-S and DeiT-Ti backbones.}
\label{tab:deit_s_ti_compression}
\end{table}

\begin{table*}[t]
\setlength{\tabcolsep}{3.53pt}
\begin{tabular}{lcccccccccccccccccc}
\toprule[1.5pt] 
\multirow{2}{*}{Method} & \multicolumn{3}{c}{DeiT-B (81.8)} & \multicolumn{3}{c}{DeiT-S (79.8)} & \multicolumn{3}{c}{DeiT-Ti (72.1)} & \multicolumn{3}{c}{Swin-B (83.4)} & \multicolumn{3}{c}{Swin-S (83.2)} & \multicolumn{3}{c}{Swin-Ti (81.2)}
\\
\cmidrule[0.5pt](lr){2-4}\cmidrule[0.5pt](lr){5-7}\cmidrule[0.5pt](lr){8-10}\cmidrule[0.5pt](lr){11-13}\cmidrule[0.5pt](lr){14-16}\cmidrule[0.5pt](lr){17-19} & 1:4 & 2:4 & 3:4 & 1:4 & 2:4 & 3:4 & 1:4 & 2:4 & 3:4 & 1:4 & 2:4 & 3:4 & 1:4 & 2:4 & 3:4 & 1:4 & 2:4 & 3:4
\\
\midrule[0.5pt] 

Random & 0.9 & 24.4 & 74.0 & 0.7 & 8.9 & 59.7 & 0.8 & 6.9 & 43.3 & 1.4 & 20.7 & 74.4 & 1.5 & 36.7 & 76.2 & 0.9 & 14.8 & 67.3 \\
Magnitude & 1.7 & 29.2 & 71.9 & 1.2 & 13.3 & 55.6 & 0.8 & 4.8 & 32.5 & 4.1 & 45.2 & 75.6 & 3.9 & 47.8 & 76.8 & 1.2 & 20.3 & 65.2 \\
Taylor FO & 6.0 & 52.4 & 78.1 & 3.9 & 39.4 & 74.8 & 3.8 & 32.7 & 63.5 & 15.4 & 65.6 & 80.9 & 11.7 & 64.8 & 80.1 & 13.6 & 60.0 & 77.0 \\
Taylor SO & 6.0 & 52.5 & 78.0 & 4.0 & 39.3 & 74.8 & 3.8 & 32.7 & 63.5 & 15.4 & 65.6 & 80.9 & 11.6 & 64.7 & 80.1 & 13.6 & 60.0 & 77.0 \\
Hessian & 5.1 & 52.0 & 78.2 & 3.8 & 48.2 & 74.8 & 2.8 & 32.2 & 63.1 & 15.5 & 64.9 & 80.7 & 11.4 & 63.8 & 80.1 & 14.3 & 60.7 & 77.2 \\
\rowcolor[HTML]{C6E2FF}WPAC (Ours) & \textbf{41.8} & \textbf{76.9} & \textbf{81.2} & \textbf{31.2} & \textbf{72.9} & \textbf{78.6} & \textbf{22.1} & \textbf{61.3} & \textbf{69.1} & \textbf{39.9} & \textbf{76.9} & \textbf{82.4} & \textbf{40.7} & \textbf{76.3} & \textbf{81.9} & \textbf{36.0} & \textbf{72.3} & \textbf{79.4} \\

\bottomrule[1.5pt] 
\end{tabular}
\caption{Comparison of direct evaluation results after pruning attention modules at different sparsity levels using various importance criteria. ``Random'' denotes random pruning; ``FO'' and ``SO'' refer to first- and second-order approximations.}
\label{tab:pruning_criteria}
\end{table*}

\begin{table}[t]
\setlength{\tabcolsep}{3.1pt}
\begin{tabular}{lccccc}
\toprule[1.5pt] 
Method & MACs & Params & Samples & Epochs & Top-1
\\
\midrule[0.5pt] 

\color{gray}DeiT-B & \color{gray}16.88 G & \color{gray}86.57 M & \color{gray}- & \color{gray}- & \color{gray}81.80 \\
\midrule[0.5pt] 

\rowcolor[HTML]{D3D3D3}\multicolumn{6}{c}{\textit{Network Compression}} \\
RePaViT & 12.70 G & 65.30 M & 1.2 M & 300 & 81.40 \\ 
IA-RED$^2$ & 11.80 G & - & 1.2 M & 90 & 80.30 \\ 
CP-ViT & 11.70 G & - & 1.2 M & 30 & 80.31 \\

WDPruning & 11.00 G & 60.60 M & 1.2 M & 100 & 81.09 \\
WDPruning & 10.80 G & 59.40 M & 1.2 M & 100 & 80.85 \\
RePaViT & 10.60 G & 51.10 M & 1.2 M & 300 & 81.30 \\ 
\rowcolor[HTML]{C6E2FF}SKD (Ours) & 10.57 G & 54.55 M & 1.2 M & 30 & \textbf{81.45} \\
WDPruning & 9.90 G & 55.30 M & 1.2 M & 100 & 80.76 \\
LPViT & 8.80 G & - & 1.2 M & 300 & 80.81 \\
X-Pruner & 8.50 G & - & 1.2 M & 130 & 81.02 \\
\rowcolor[HTML]{C6E2FF}SKD (Ours) & 8.47 G & 43.92 M & 1.2 M & 30 & \textbf{81.24} \\
UVC & 8.00 G & - & 1.2 M & 300 & 80.57 \\
LPViT & 7.92 G & - & 1.2 M & 300 & 80.55 \\

\rowcolor[HTML]{C6E2FF}SKD (Ours) & 7.89 G & 40.96 M & 1.2 M & 30 & \textbf{80.89} \\



\midrule[0.5pt] 

\rowcolor[HTML]{D3D3D3}\multicolumn{6}{c}{\textit{Few-shot Compression}} \\
PRACTISE & 14.43 G & - & 500 & 2000 & 79.30 \\
DC-DeiT & 14.09 G & - & 500 & 4000 & 81.26 \\
\rowcolor[HTML]{C6E2FF}SKD (Ours) & 13.42 G & 69.01 M & 0 & 0 & \textbf{81.42}
\\

\bottomrule[1.5pt] 
\end{tabular}
\caption{Performance comparison with network compression methods on ImageNet-1k. All methods adopt DeiT-B as the base backbone. ``Samples'' and ``Epochs'' indicate the number of samples and epochs used for fine-tuning, respectively.}
\label{tab:deit_b_compression}
\end{table}

\begin{table}[t]
\setlength{\tabcolsep}{4.2pt}
\begin{tabular}{lccccccc}
\toprule[1.5pt] 
Method & \rotatebox[origin=c]{90}{\makecell{Flowers-102}} & \rotatebox[origin=c]{90}{\makecell{CUB-200-2011}} & \rotatebox[origin=c]{90}{\makecell{Stanford Cars}} & \rotatebox[origin=c]{90}{\makecell{CIFAR-10}} & \rotatebox[origin=c]{90}{\makecell{CIFAR-100}} & \rotatebox[origin=c]{90}{\makecell{Food-101}} & \rotatebox[origin=c]{90}{\makecell{\textit{Average}}}
\\
\midrule[0.5pt] 

Albert & 94.1 & 72.4 & 87.2 & 96.5 & 78.5 & 83.0 & \textit{85.3} \\
LiGO & 95.9 & 74.8 & 87.9 & 96.9 & 81.3 & 84.0 & \textit{86.8} \\
Heur-LG & 69.1 & 48.0 & 51.2 & 95.1 & 72.8 & 76.8 & \textit{68.8} \\
Auto-LG & 96.4 & 75.1 & 88.2 & 97.3 & 81.0 & 84.6 & \textit{87.1} \\
TLEG & 93.7 & 72.6 & 87.2 & 97.2 & 80.2 & 84.9 & \textit{86.0} \\
WAVE & \textbf{96.9} & 78.1 & 89.4 & 97.4 & 83.2 & 85.5 & \textit{88.4} \\
\rowcolor[HTML]{C6E2FF}SKD (Ours) & 94.1 & \textbf{81.5} & \textbf{90.5} & \textbf{98.0} & \textbf{86.6} & \textbf{89.2} & \textbf{\textit{90.0}} \\

\bottomrule[1.5pt] 
\end{tabular}
\caption{Comparison with network expansion methods on downstream datasets, using a DeiT-S backbone with half the original size. ``\textit{Average}'' denotes the average accuracy.}
\label{tab:transferability}
\end{table}

\paragraph{Comparison with Network Expansion Methods.}
To efficiently obtain models of varying sizes, methods like Albert \cite{lanalbert}, LiGO \cite{wanglearning}, and Heur-LG \cite{wang2022learngene} use weight decomposition and reuse to initialize different-sized models from pretrained weights. Auto-LG \cite{wang2023learngene} automates layer selection for inheritance. TLEG \cite{xia2024transformer} applies cross-layer weight fusion, SWS \cite{xia2024exploring} employs stage-wise weight sharing, and WAVE \cite{feng2025wave} uses weight templates to build models of varying sizes.
Table \ref{tab:network_expansion} compares our method with these approaches. Our method requires no additional teacher model for distillation, and the extracted sub-networks achieve better performance without any fine-tuning.

\paragraph{Comparison with Network Compression Methods.}
The construction of scalable models remains an underexplored area. Therefore, we compare our method with a range of state-of-the-art model compression approaches, including RePaViT \cite{xurepavit}, IA-RED$^2$ \cite{pan2021ia}, CP-ViT \cite{song2022cp}, WDPruning \cite{yu2022width}, LPViT \cite{xu2024lpvit}, X-Pruner \cite{yu2023x}, UVC \cite{yuunified}, SPViT \cite{he2024pruning}, as well as few-shot compression methods such as PRACTISE \cite{wang2023practical} and DC-ViT \cite{zhang2024dense}.
Table \ref{tab:deit_s_ti_compression} and Table \ref{tab:deit_b_compression} present the comparison results. Experimental findings demonstrate that our method can produce more compact models with comparable or even superior performance, while requiring significantly fewer fine-tuning resources.

\paragraph{Comparison with Importance Criteria.}
Many existing pruning methods rely on well-designed importance criteria to evaluate the significance of each weight. However, existing importance criteria for structured pruning evaluate each dimension independently, without the ability to jointly consider or integrate information across multiple dimensions.
Table \ref{tab:pruning_criteria} compares our proposed WPAC with these criteria. The results demonstrate that WPAC effectively integrates information across dimensions, enabling more reliable pruning and benefiting subsequent model compression.

\paragraph{Comparison of Transferability.}
Table \ref{tab:transferability} presents the transferability of the sub-networks produced by our method on downstream tasks. We evaluate the transferability of the models on the following datasets: Flowers-102 \cite{nilsback2008automated}, CUB-200-2011 \cite{WahCUB_200_2011}, Stanford Cars \cite{krause20133d}, CIFAR-10 \cite{krizhevsky2009learning}, CIFAR-100 \cite{krizhevsky2009learning}, and Food-101 \cite{bossard2014food}. Compared to other efficient approaches for obtaining models of different sizes, the models generated by our method exhibit superior transfer performance across various datasets.

\subsection{Ablation and Analysis}
\begin{figure}[t]
    \centering
    \includegraphics[width=\linewidth]{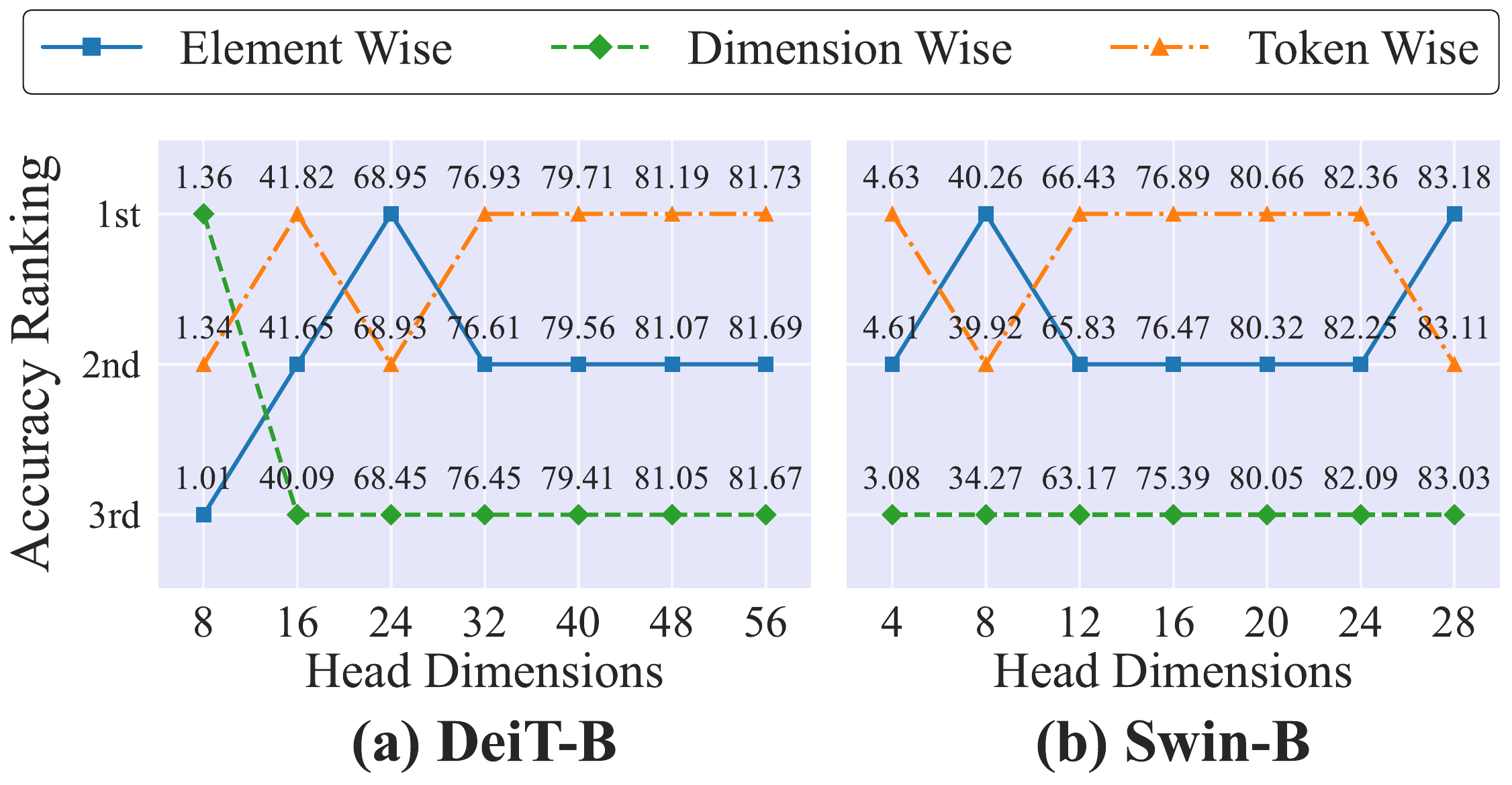}
    \caption{Comparison of weighting strategies with different granularities across varying numbers of retained dimensions.}
    \label{fig:weighting_granularities}
\end{figure}

\begin{table}[t]
\setlength{\tabcolsep}{4.9pt}
\begin{tabular}{ccccc}
\toprule[1.5pt] 
Tokens & Weight & DeiT-B & DeiT-S & DeiT-Ti
\\
\midrule[0.5pt] 

Class & 1.0 & 65.1$\pm$0.15 & 46.2$\pm$0.20 & 30.8$\pm$0.28 \\
Rand(1) & 1.0 & 73.3$\pm$0.19 & 64.5$\pm$0.54 & 47.4$\pm$1.38 \\
Rand(2) & 1.0 & 75.7$\pm$0.17 & 71.0$\pm$0.11 & 58.7$\pm$0.10 \\
Rand(5) & 1.0 & Ill-Cond. & 71.4$\pm$0.07 & 59.4$\pm$0.28 \\
All & 1.0 & Ill-Cond. & Ill-Cond. & Ill-Cond. \\
All & 0.01 & 76.1$\pm$0.06 & 72.2$\pm$0.07 & 60.0$\pm$0.05 \\
\rowcolor[HTML]{C6E2FF}All & Imp. & \textbf{76.9}$\pm$0.05 & \textbf{72.9}$\pm$0.07 & \textbf{61.2}$\pm$0.07 \\

\bottomrule[1.5pt] 
\end{tabular}
\caption{WPAC results for different token selection and weighting strategies when retaining 50\% dimensions. ``Class'' uses the class token, ``Rand($x$)'' randomly selects $x$ tokens, and ``All'' uses all tokens per sample. ``Imp.'': token-wise importance weighting. ``Ill-Cond.'': eigen decomposition failure.}
\label{tab:token_sampling}
\end{table}

\begin{figure}[t]
    \centering
    \begin{subfigure}{0.49\linewidth}
        \centering
        \includegraphics[width=\linewidth]{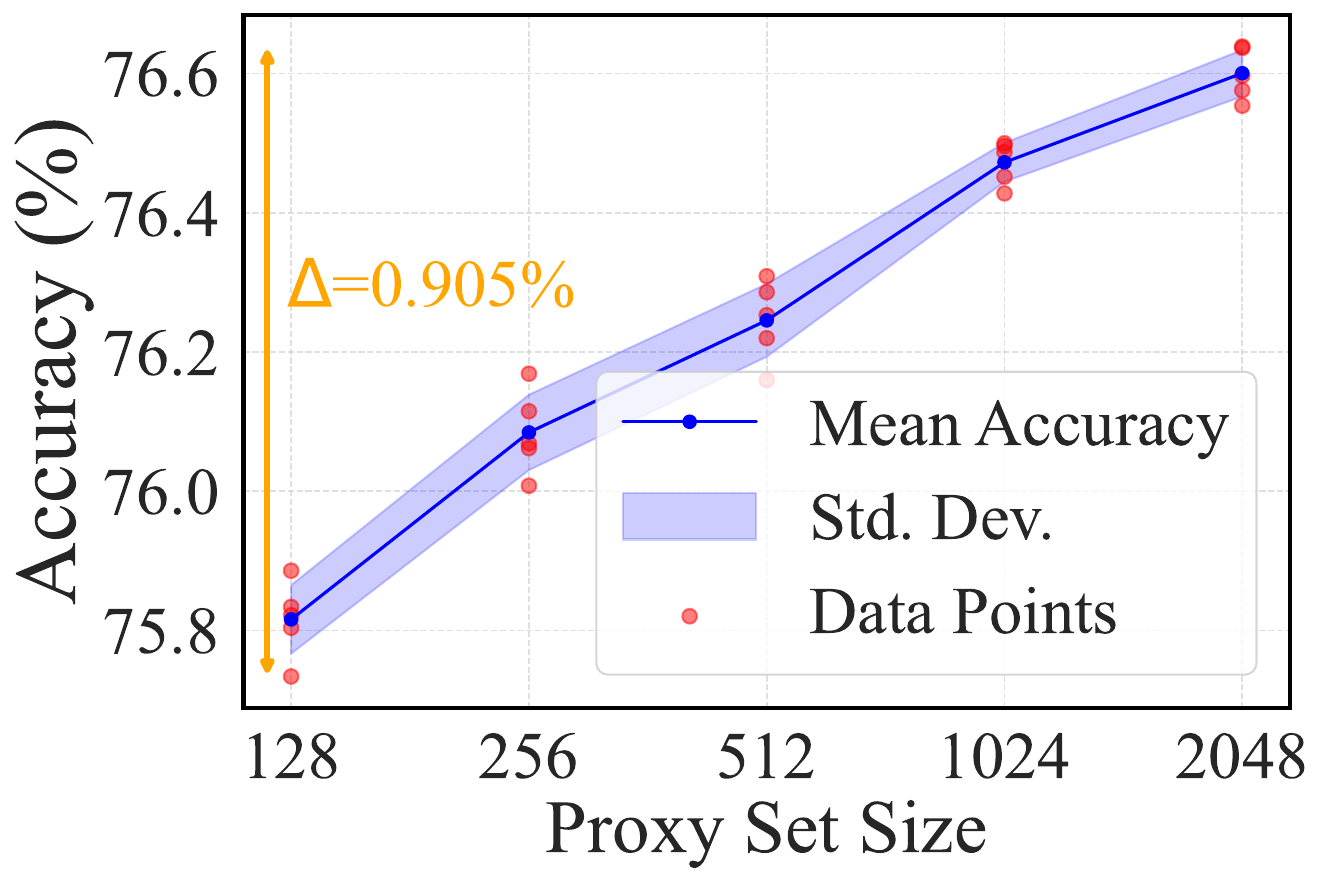}
        \caption{DeiT-B}
    \end{subfigure}
    \hfill
    \begin{subfigure}{0.49\linewidth}
        \centering
        \includegraphics[width=\linewidth]{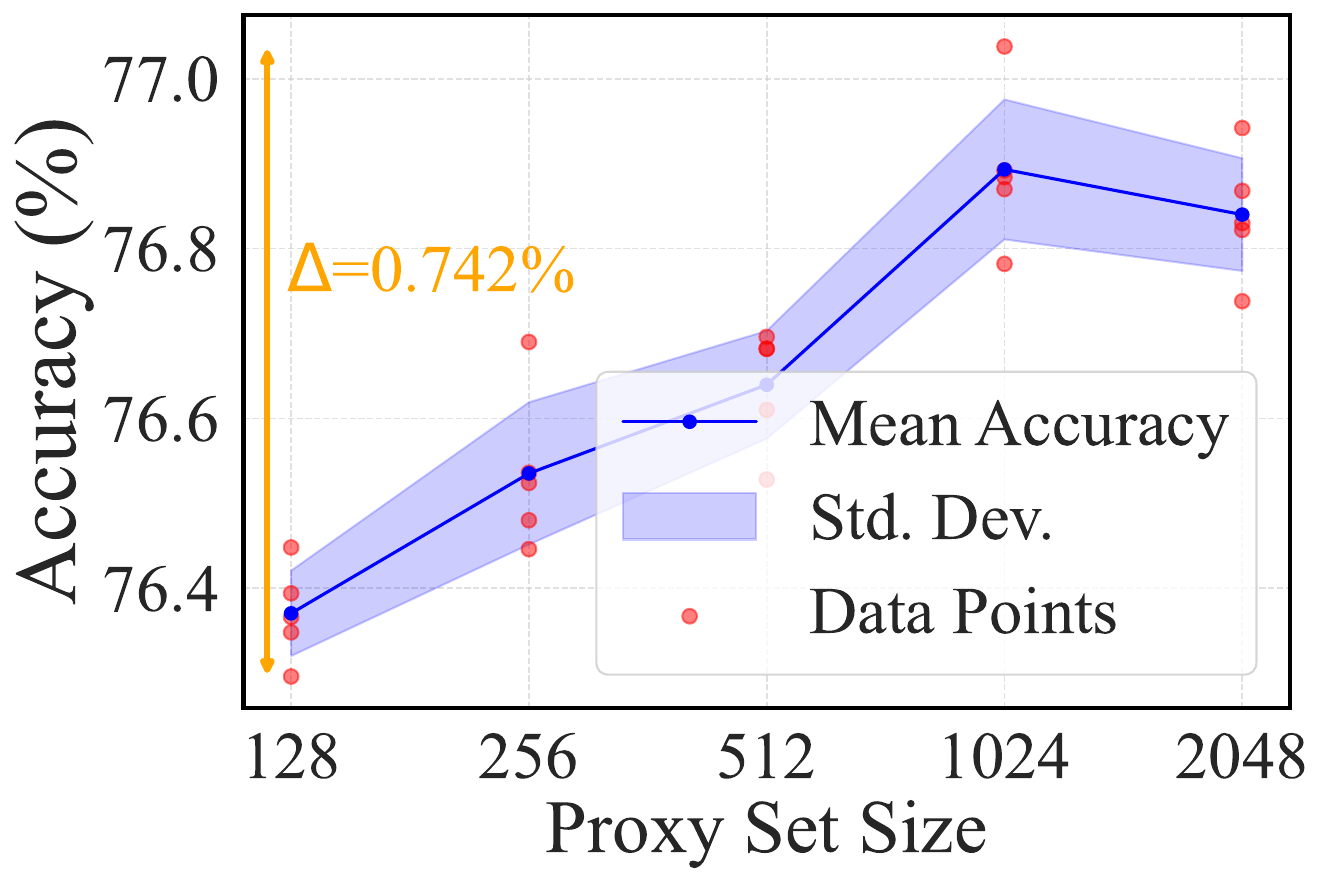}
        \caption{Swin-B}
    \end{subfigure}
    \caption{Impact of proxy set size on WPAC performance. Each size is evaluated using 5 random seeds, and the results reflect performance with head dimensions reduced by half.}
    \label{fig:proxy_size}
\end{figure}

\begin{table}[t]
\setlength{\tabcolsep}{2.8pt}
\begin{tabular}{lcccccccc}
\toprule[1.5pt] 
\multirow{2}{*}{Method} & \multicolumn{4}{c}{DeiT-S (79.8)} & \multicolumn{4}{c}{DeiT-Ti (72.1)}
\\
\cmidrule[0.5pt](lr){2-5}\cmidrule[0.5pt](lr){6-9} & 4:12 & 6:12 & 8:12 & 10:12 & 4:12 & 6:12 & 8:12 & 10:12
\\
\midrule[0.5pt] 

Baseline & 1.2 & 8.9 & 39.1 & 71.2 & 1.4 & 6.9 & 25.8 & 56.2 \\
B+CD & 7.3 & 20.0 & 34.4 & 46.5 & 2.0 & 8.8 & 23.1 & 36.3 \\
B+WCD & 34.2 & 54.4 & 64.7 & 73.8 & 29.4 & 42.2 & 49.1 & 63.8 \\
B+LD & 39.7 & 60.2 & 68.6 & 74.1 & 34.5 & 45.4 & 57.1 & 62.5 \\
\rowcolor[HTML]{C6E2FF}B+PIAD & \textbf{70.6} & \textbf{76.2} & \textbf{78.2} & \textbf{79.0} & \textbf{61.4} & \textbf{65.8} & \textbf{68.6} & \textbf{70.0} \\

\bottomrule[1.5pt] 
\end{tabular}
\caption{Direct evaluation of sub-networks of varying sizes extracted from networks trained with different dropout strategies. ``$x$:12'': the subnetwork-to-original size ratio. ``CD'': channel dropout. ``WCD'': weighted CD. ``LD'': layerdrop.}
\label{tab:ablation_piad}
\end{table}

\paragraph{Analysis of Weighting Strategies with Varying Granularities in WPAC.}
Given the element-wise score $\Theta_{\text{TE}} \in \mathbb{R}^{n \times d}$ obtained via Taylor expansion, aggregating along different axes yields token-wise and dimension-wise importance scores. Figure \ref{fig:weighting_granularities} presents the results and rankings of the three weighting strategies. Experimental results show that token-wise weighting is more effective at preserving information and provides greater robustness compared to the others.

\paragraph{Analysis of Per-Sample Token Sampling Methods and Weighting Effectiveness in WPAC.}
Selecting appropriate tokens for PCA is crucial for preserving essential features. Table \ref{tab:token_sampling} presents results for different token sampling strategies on each sample in the proxy set: using only the class token, randomly selecting a subset of tokens, using all tokens, and applying different weighting schemes. Relying solely on the class token limits the ability to recover information from other tokens, leading to the poorest performance. Randomly sampling several tokens introduces more sample information into PCA, but using too many tokens can result in an ill-conditioned covariance matrix, making eigen decomposition infeasible. In contrast, our WPAC method utilizes all tokens with importance-based weighting, achieving the best results.

\paragraph{Analysis of the Impact of Proxy Dataset Size in WPAC.}
Figure \ref{fig:proxy_size} illustrates the results of applying PCA with proxy sets of varying sizes. The experiments demonstrate that even a small number of samples is sufficient to obtain an accurate principal component projection. This allows the pretrained parameters to be effectively projected, concentrating the knowledge into a reduced set of transformed dimensions.

\paragraph{Effectiveness of the PIAD.}
Table \ref{tab:ablation_piad} shows the impact of different dropout strategies on sub-network performance.
Channel Dropout \cite{tompson2015efficient} and LayerDrop \cite{fanreducing} tend to favor sub-networks with sparsity close to the preset dropout rate and offer limited sub-network diversity.
Weighted Channel Dropout \cite{hou2019weighted} further concentrates knowledge into important dimensions by assessing their significance.
However, all these methods suffer from uneven sub-network sampling across sizes.
In contrast, our PIAD method effectively condenses knowledge into fewer parameters and forms a hierarchical structure, enabling incrementally expandable networks.

\section{Conclusion}
We propose constructing a Stratified Knowledge-Density (SKD) super-network for ViTs, enabling efficient extraction of sub-networks with maximal knowledge retention to meet diverse deployment needs. WPAC condenses knowledge into key dimensions, laying a solid foundation for compression, while PIAD further enhances knowledge stratification across weight groups to form a stratified super-network.

\section{Acknowledgments}
This research was supported by the Jiangsu Science Foundation (BG2024036, BK20243012), the National Science Foundation of China (62125602, U24A20324, 92464301),  CAAI-Lenovo Blue Sky Research Fund, and the Fundamental Research Funds for the Central Universities (2242025K30024).

\bibliography{aaai2026}

\end{document}